\documentclass[11pt,a4paper]{article}
\usepackage[hyperref]{naaclhlt2018}
\usepackage{times}
\usepackage{latexsym}

\usepackage{url}

\usepackage{amsfonts}       %
\usepackage{nicefrac}       %
\usepackage{microtype}      %
\usepackage{pgf}
\usepackage{apalike}
\usepackage{times}
\usepackage{latexsym}
\usepackage{amsmath}
\usepackage{amssymb}
\usepackage{graphicx}
\usepackage{color}
\usepackage{paralist}
\usepackage{tikz}
\usetikzlibrary{bayesnet}
\usepackage{xspace}
\usepackage{subcaption}
\usepackage{physics}
\usepackage{comment}

\newcommand{\angbrack}[1]{\ensuremath{\langle #1 \rangle}}

\newcommand{\reffig}[1]{Figure~\ref{fig:#1}}

\newcommand{\dx}{\ensuremath{{d_{\texttt{x}}}}}

\newcommand{\vx}{\ensuremath{{v_{\texttt{x}}}}}
\newcommand{\vy}{\ensuremath{{v_{\texttt{y}}}}}
\newcommand{\txt}[1]{{\ensuremath{\text{#1}}}}
\DeclareMathOperator{\KL}{KL}
\DeclareMathOperator{\diag}{diag}
\DeclareMathOperator{\Cat}{Cat}

\DeclareMathOperator{\softmax}{softmax}
\DeclareMathOperator{\softplus}{softplus}

\newcommand{\Lx}{\texttt{L1}\xspace}
\newcommand{\Ly}{\texttt{L2}\xspace}

\usepackage{pgfplotstable}
\usepackage{booktabs}
\usepackage{array}
\usepackage{colortbl}

\aclfinalcopy %

\title{Deep Generative Model for Joint Alignment and Word Representation\thanks{Code available from \url{https://github.com/uva-slpl/embedalign}}}

\author{Miguel Rios \qquad Wilker Aziz  \qquad Khalil Sima'an \thanks{MR and WA contributed equally.} \\
  Institute for Logic, Language, and Computation\\
  University of Amsterdam \\
  {\tt \{m.riosgaona, w.aziz, k.simaan\}@uva.nl}  }

\date{}

\renewcommand\footnotemark{}

\begin{document}

\maketitle

\begin{abstract}

This work exploits translation data as a source of semantically relevant learning signal for models of word representation. 
In particular, we exploit equivalence through translation as a form of distributed context and jointly learn how to embed and align with a deep generative model. 
Our \textsc{EmbedAlign} model embeds words in their complete observed context and learns by marginalisation of latent lexical alignments.
Besides, it embeds words as posterior probability densities, rather than point estimates, which allows us to compare words in context using a measure of overlap between distributions (e.g. $\KL$ divergence).  
We investigate our model's performance on a range of lexical semantics tasks achieving competitive results on several standard benchmarks 
 including natural language inference, paraphrasing, and text similarity.

\end{abstract}
\section{Introduction}

Natural language processing applications %
 often count on the availability of word representations trained on large textual data as a means to alleviate problems such as data sparsity and lack of linguistic resources \citep{CollobertEtAl2011NLP,SocherEtAl2011SSEA,TuEtAl2017RepL4NLP,BowmanEtAlNLI}.

Traditional approaches to inducing word representations circumvent the need for explicit semantic annotation by capitalising on some form of indirect semantic supervision.
A typical example is to fit a binary classifier to detect whether or not a target word is likely to co-occur with neighbouring words \citep{Mikolov+2013:DR}.
If the binary classifier represents a word as a continuous vector, that vector will be trained to be discriminative of the contexts it co-occurs with, and thus words in similar contexts will have similar representations.

The underlying assumption is that context (e.g. neighbouring words) stands for the meaning of the target word \citep{Harris:1954,Firth:1957}.
The success of this \emph{distributional hypothesis} hinges on the definition of context and different models are based on different definitions. 
Importantly, the nature of the context determines the range of linguistic properties the representations may capture \citep{levy-goldberg14regularities}.
For example, \citet{LevyEtAl2014DWE} propose to use syntactic context derived from dependency parses. They show that their representations are much more discriminative of syntactic function than models based on immediate neighbourhood \citep{Mikolov+2013:DR}.
%

\begin{comment}
On the one hand, binary classification is rather efficient. 
On the other hand, the strategy for generating labelled data is slightly artificial---words far from the target word still have co-occurred with it even though we take them as negative evidence.
In scenarios where the immediate neighbourhood is less discriminative or shows high variability, this strategy becomes less effective---which can happen for languages with productive morphology \citep{NOT-SURE}. 
\end{comment}
%

In this work, we take lexical translation as indirect semantic supervision \citep{DiabEtAl2002WSD}. 
Effectively we make two assumptions. First, that every word has a foreign equivalent that stands for its meaning. Second, that we can find this equivalent in translation data through lexical alignments.\footnote{These assumptions are not new to the community, but in this work they lead to a novel model which reaches more applications. \S\ref{sec:related} expands on the relation to other uses of bilingual data for word representation.}
For that we induce both a latent mapping between words in a bilingual sentence pair and distributions over latent word representations. 

To summarise our contributions: 
\begin{itemize}\setlength\itemsep{0em}
	\item we model a joint distribution over sentence pairs that generates data from latent word representations and latent lexical alignments;
	\item we embed words in context mining positive correlations from translation data;
	\item we find that foreign observations are necessary for generative training, but test time predictions can be made monolingually;
	\item we apply our model to a range of semantic natural language processing tasks showing its usefulness.
\end{itemize}
%
%
%
%
%

\begin{comment} %

\citet{thater+11} propose a method that represents word meaning in context by modifying the embeddings according to the neighbouring syntactic context.
\citet{Melamud+15:naacl} propose to combine word embeddings with the context for meaning representation. 
In contrast, \citet{Kocisky+14:vecibm2} use parallel data for inducing a meaning representation as well as word alignments to capture cross-lingual semantic information about the context. 
 In this work, we use lexical translation as learning signal, where equivalence through translation is a proxy to meaning equivalence for inducing word representations.
%
%
\end{comment}

%

%

%
%

\section{\label{sec:embed-align}\textsc{EmbedAlign}}

In a nutshell, we model a distribution over pairs of sentences expressed in two languages, namely, a language of interest \Lx, and an auxiliary language \Ly which our model uses to mine some learning signal.
Our model, \textsc{EmbedAlign}, is governed by a simple generative story:
\begin{enumerate}\setlength\itemsep{0em}
	\item sample a length $m$ for a sentence in \Lx and a length $n$ for a sentence in \Ly;
	\item generate a sequence $z_1, \ldots, z_m$ of $d$-dimensional random embeddings by sampling independently from a standard Gaussian prior;
	\item generate a word observation $x_i$ in the vocabulary of \Lx conditioned on the random embedding $z_i$;
	\item generate a sequence $a_i, \ldots, a_n$ of $n$ random alignments---each maps from a position $a_j$ in $x_1^m$ to a position $j$ in the \Ly sentence;
	\item finally, generate an observation $y_j$ in the vocabulary of \Ly conditioned on the random embedding $z_{a_j}$ that stands for $x_{a_j}$.
\end{enumerate}
The model is parameterised by neural networks and parameters are estimated to maximise a lowerbound on log-likelihood of joint observations.
In the following, we present the model formally (\S\ref{sec:pgm}), discuss efficient training (\S\ref{sec:barber}), and concrete architectures (\S\ref{sec:arch}). %

\subsection{\label{sec:pgm}Probabilistic model}

\begin{figure}[t]
\centering
\begin{tikzpicture}
\node[obs]	                   (x)		{$ x $};
\node[obs, below = of x]	       (y)		{$ y $};
\node[latent, left = of x]		(z)		{$ z $};
\node[latent, left = of y]		(a)		{$ a $};
\node[right = of x] (theta) {$\theta$};

\edge{a}{y};
\edge{z}{x};
\edge{z}{y};
\edge{theta}{y,x};

\plate {L1-sentence} {(x)(z)} {$ m $};
\plate {L2-sentence} {(y)(a)} {$ n $};
\plate {corpus} {(L1-sentence) (L2-sentence)} {$|\mathcal B|$};
\end{tikzpicture}
\caption{\label{fig:model} A sequence $x_1^m$ is generated conditioned on a sequence of random embeddings $z_1^m$; generating the foreign sequence $y_1^n$ further requires latent lexical alignments $a_1^n$.} %
\end{figure}
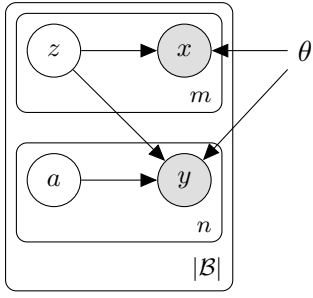

\paragraph{Notation} We use block capitals (e.g. $X$) for random variables, lowercase letters (e.g. $x$) for assignments, and the shorthand $X_1^m$ for a sequence $X_1, \ldots, X_m$. Boldface letters are reserved for deterministic vectors (e.g. $\mathbf v$) and matrices (e.g. $\mathbf W$). Finally, $\mathbb E[f(Z); \alpha]$ denotes the expected value of $f(z)$ under a density $q(z| \alpha)$.

We model a joint distribution over bilingual parallel data, i.e., \Lx--\Ly sentence pairs. %
An observation is a pair of random sequences $\angbrack{X_1^m, Y_1^n}$, where a random variable $X$ ($Y$) takes on values in the vocabulary of \Lx (\Ly).
For ease of exposition, the length $m$ ($n$) of each sequence is assumed observed throughout.
The \Lx sentence is generated one word at a time from a random sequence of latent embeddings $Z_1^m$, each $Z$ taking on values in $\mathbb R^d$.
The \Ly sentence is generated one word at a time given a random sequence of latent alignments $A_1^n$, where $A_j \in \{1, \dotsc, m\}$ is the position in the \Lx sentence to which $y_j$ aligns.\footnote{We pad \Lx sentences with \textsc{Null} to account for untranslatable \Ly words \citep{Brown+1993:smt}. Instead, \citet{Schulz+2016:BIBM2Z} generate untranslatable words from \Ly context---an alternative we leave for future work.}

For $i \in \{1, \ldots, m\}$ and $j \in \{1, \ldots, n\}$ the generative story is
\begin{subequations}
\begin{align}
 Z_i &\sim \mathcal N(\mathbf 0, \mathbf I) \\ %
 X_i|z_i &\sim \Cat(f(z_i; \theta)) \\ %
A_j|m &\sim \mathcal U(1/m) \\ %
Y_j|z_1^m, a_j &\sim \Cat(g(z_{a_j}; \theta))  %
\end{align}
\end{subequations}
and  \reffig{model} is a graphical depiction of our model.  %
We map from latent embeddings to %
 categorical distributions over either vocabulary using a neural network whose parameters are deterministic and collectively denote by $\theta$ (the \emph{generative parameters}). 
The marginal likelihood of a sentence pair is shown in Equation (\ref{eq:marginal}). %
\begin{equation}\label{eq:marginal}
\begin{aligned}
P_\theta(x_1^m, y_1^n|m, n) = \int p(z_1^m) \prod_{i=1}^m P_\theta(x_i|z_i) \\
 \times \prod_{j=1}^n \sum_{a_j=1}^m P(a_j|m) P_\theta(y_j|z_{a_j}) \dd z_1^m
\end{aligned}
\end{equation}

Due to the conditional independences of our model, it is trivial to marginalise lexical alignments for any given latent embeddings $z_1^m$, but marginalising the embeddings themselves is intractable. %
Thus, we employ amortised mean field variational inference using the inference model
\begin{equation}\label{eq:qz}
\begin{aligned}
q_\phi(z_1^m|x_1^m) \triangleq  \prod_{i=1}^m \mathcal N(z_i|\mathbf u_i, \diag(\mathbf s_i \odot \mathbf s_i))
\end{aligned}
\end{equation}
where each factor is a diagonal Gaussian.
We map from $x_1^m$ to a sequence $\mathbf u_1^m$ of independent posterior mean (or location) vectors, where $\mathbf u_i \triangleq \mu(\mathbf h_i; \phi)$, as well as a sequence $\mathbf s_1^m$ of independent standard deviation (or scale) vectors, where $\mathbf s_i   \triangleq \sigma(\mathbf h_i; \phi)$, and $\mathbf h_1^m = \mathrm{enc}(x_1^m; \phi)$ is a deterministic encoding of the \Lx sequence (we discuss concrete architectures in \S\ref{sec:arch}). 
All mappings are realised by neural networks whose parameters are collectively denoted by $\phi$ (the \emph{variational parameters}). 
Note that we choose to approximate the posterior without conditioning on $y_1^n$. This allows us to use the inference model for \emph{monolingual} prediction in absence of \Ly data.

Variational $\phi$ and generative $\theta$ parameters are jointly point-estimated to attain a local optimum of the evidence lowerbound \citep{Jordan+1999:VI}:
\begin{equation}\label{eq:ELBO}
\begin{aligned}
\log P_\theta(x_1^m, y_1^n|m, n) \ge \\ %
 \sum_{i=1}^m \mathbb E \left[ \log P_\theta(x_i|Z_i); \mathbf u_i, \mathbf s_i \right] \\
 + \sum_{j=1}^n \mathbb E \left[  \log \sum_{a_j=1}^m P(a_j|m) P_\theta(y_j|Z_{a_j}); \mathbf u_1^m, \mathbf s_1^m \right] \\
 - \sum_{i=1}^m \KL\left[\mathcal N(\mathbf u_i, \diag(\mathbf s_i \odot \mathbf s_i)) || \mathcal N(\mathbf 0, \mathbf I)\right] \quad .
\end{aligned}
\end{equation}
The variational family is location-scale, thus we can rely on stochastic optimisation \citep{Robbins+1951:SA} and automatic differentiation \citep{BaydinEtAl2015AD} with reparameterised gradient estimates \citep{Kingma+2014:VAE,Rezende+14:DGM,TitsiasEtAl2014doubly}. Moreover, because the Gaussian density is an exponential family, the $\KL$ terms in (\ref{eq:ELBO}) are available in closed-form \citep[Appendix B]{Kingma+2014:VAE}. 

\subsection{\label{sec:barber}Efficient training}

The likelihood terms in the ELBO (\ref{eq:ELBO}) require evaluating two $\softmax$ layers over rather large vocabularies.
This makes training prohibitively slow and calls for efficient approximation.
We employ an approximation proposed by \citet{BotevEtAl2017CSS} termed \emph{complementary sum sampling} (CSS), which we review in this section.

Consider the likelihood term $\log P(X=\mathsf x|z)$ that scores an observation $\mathsf x$ given a sampled embedding $z$---we use serif font $\mathsf x$ to distinguish a particular observation from an arbitrary event $x \in \mathcal X$ in the support. 
The exact class probability 
\begin{equation}
P(X=\mathsf x|z) = \frac{\exp(u(z, \mathsf x))}{\sum_{x \in \mathcal X} \exp(u(z, x))}
\end{equation}
requires a normalisation over the complete support.
CSS works by splitting the support into two sets, a set $\mathcal C$ that is explicitly summed over and must include the \emph{positive} class $\mathsf x$, and another set $\mathcal N$ that is a subset of the complement set $\mathcal X \setminus C$. 
We obtain an estimate for the normaliser
\begin{equation}\label{eq:cssnormaliser}
\sum_{x \in \mathcal C} \exp(u(z, x)) + \sum_{x \in \mathcal N} \kappa(x) \exp(u(z, x))
\end{equation}
 by importance- or Bernoulli-sampling from the support using a proposal distribution $Q(X)$, where $\kappa(x)$ corrects for bias as $\mathcal N$ tends to the entire complement set.
 In this paper, we design $\mathcal C$ and $\mathcal N$ per training mini-batch: we take $\mathcal C$ to consist of all unique words in a mini-batch of training samples and $\mathcal N$ to consist of $10^3$ negative classes uniformly sampled from the complement set $\mathcal X \setminus \mathcal C$, in which case $\kappa(x) = 10^{-3}|\mathcal X \setminus \mathcal C|$.\footnote{We sample uniformly from the complement set until we have $10^3$ unique classes. We realise this operation outside the computation graph providing $\mathcal C$ and $\mathcal N$ as inputs to each training iteration, but a GPU-based solution is also possible.}

CSS makes it particularly easy to approximate likelihood terms such as those with respect to \Ly in Equation (\ref{eq:ELBO}).
Because those terms depend on a marginalisation over alignments, an approximation must give support to all words in the sequence $y_1^n$. With CSS this is extremely simple, we just need to make sure all unique words in $y_1^n$ are in the set $\mathcal C$---which our mini-batch procedure does guarantee.
\citet{BotevEtAl2017CSS} show that CSS is rather stable and superior to the most popular $\softmax$ approximations.
Besides being simple to implement, CSS also addresses a few problems with other approximations. To name a few: unlike importance sampling approximations, CSS converges to the exact $\softmax$ with bounded computation (it takes as many samples as there are classes). Unlike hierarchical $\softmax$, CSS only affects training, that is, at test time we simply use the entire support instead of the approximation.

Without a $\softmax$ approximation, inference for our model would take time proportional to $O(m \times \vx + m \times \vy + m \times n)$ where $\vx$ ($\vy$) corresponds to the size of the vocabulary of \Lx (\Ly). The first term ($m \times \vx$) corresponds to projecting from $m$ latent embeddings to $m$ categorical distributions over the vocabulary of \Lx. The second term ($m \times \vy$) corresponds to projecting the same $m$ latent embeddings to $m$ categorical distributions over the vocabulary of \Ly. Finally, the third term ($m \times n$) is due to marginalisation of alignments. 
Note, however, that with the CSS approximation we drop the dependency on vocabulary sizes (as the combined sizes of $\mathcal C$ and $\mathcal N$ is an independent constant).  
Moreover, if inference is performed on GPU, the squared term ($m \times n \approx m^2$) is amortised due to parallelism. Thus, while training our model is somewhat slower than monolingual models of word representation, which typically run in $O(m)$, it is not at all impracticably slower. 

\subsection{\label{sec:arch}Architectures}

Here we present the neural network architectures that parameterise the different generative and variational components of \S\ref{sec:pgm}.
Refer to Appendix \ref{app:architecture} for an illustration.

\paragraph{Generative model} We have two generative components, namely, a categorical distribution over the vocabulary of \Lx and another over the vocabulary of \Ly. We predict the parameter (event probabilities) of each distribution with an affine transformation of a latent embedding followed by the $\softmax$ nonlinearity to ensure normalisation: %
\begin{subequations}
\begin{align}
f(z_i; \theta) &= \softmax\left(\mathbf W_\txt{1}  z_i + \mathbf b_\txt{1}\right) \label{eq:px}\\
g(z_{a_j}; \theta) &= \softmax\left( \mathbf W_\txt{2}  z_{a_j} + \mathbf b_\txt{2} \right) \label{eq:py}
\end{align}
\end{subequations}
where $\mathbf W_\txt{1} \in \mathbb R^{\vx \times d}$, $\mathbf b_\txt{1} \in \mathbb R^\vx$,  $\mathbf W_\txt{2} \in \mathbb R^{\vy \times d}$, $\mathbf b_\txt{2} \in \mathbb R^\vy$, and $\vx$ ($\vy$) is the size of the vocabulary of \Lx (\Ly). With the approximation of \S\ref{sec:barber}, we replace the \Lx $\softmax$ layer (\ref{eq:px}) by $\exp(z_i^\top \mathbf c_x + b_x)$ normalised by the CSS estimate (\ref{eq:cssnormaliser}) at training, and similarly for the \Ly $\softmax$ layer (\ref{eq:py}). In that case, we have parameters for $\mathbf c_x, \mathbf c_y \in \mathbb R^d$---deterministic embeddings for $x$ and $y$, respectively---as well as bias terms $b_x, b_y \in \mathbb R$.

\paragraph{Inference model} We predict approximate posterior parameters using two independent transformations
\begin{subequations}\label{eq:posterior-params}
\begin{align}
\mathbf u_i      &= \mathbf M_\txt{1}  \mathbf h_i + \mathbf d_\txt{1}  \\
\mathbf s_i &= \softplus(\mathbf M_\txt{2}  \mathbf h_i + \mathbf d_\txt{2}) 
\end{align}
\end{subequations} 
of a shared representation $\mathbf h_i \in \mathbb R^\dx$ of the $i$th word in the \Lx sequence $x_1^m$---where 
$\mathbf M_\txt{1}, \mathbf M_\txt{2} \in \mathbb R^{d \times \dx}$ are projection matrices, $\mathbf d_\txt{1}, \mathbf d_\txt{2} \in \mathbb R^d$ are bias vectors, and the $\softplus$ nonlinearity ensures that standard deviations are non-negative.
To obtain the deterministic encoding $\mathbf h_1^m$, we employ two different architectures: (1) a bag-of-words (\textsc{BoW}) encoder, where $\mathbf h_i$ is a deterministic projection of $x_i$ onto $\mathbb R^\dx$; and (2) a bidirectional (\textsc{BiRNN}) encoder, where $\mathbf h_i$ is the element-wise sum of two LSTM hidden states ($i$th step) that process the sequence in opposite directions.  %
We use $128$ units for deterministic embeddings, and $100$ units for LSTMs \citep{HochreiterEtAl1997LSTM} and latent representations (i.e. $d=100$).

\section{\label{sec:exp}Experiments}

We start the section describing the data used to estimate our model's parameters as well as details about the optimiser.
The remainder of the section presents results on various benchmarks. 

\paragraph{Training data} We train our model on bilingual parallel data. 
In particular, we use parliament proceedings (Europarl-v7) \citep{koehn2005epc} from two language pairs: English-French and English-German.\footnote{The proposed model is not limited to these language pairs.}
We employed very minimal preprocessing, namely, tokenisation and lowercasing using scripts from \textsc{Moses} \citep{koehn2007moses}, and have discarded sentences longer than $50$ tokens.
Table \ref{tab:corpora} lists more information about the training data, including 
the English-French Giga web corpus \citep{bojar2014wmt} which we use in \S\ref{sec:wordsim}.\footnote{As we investigate various configurations and train every model $10$ times to inspect variance in results, we conduct most of the experiments on the more manageable Europarl.}

\begin{table}[h]\centering
\begin{tabular}{l r r}
\toprule
Corpus     & Sentence pairs & Tokens \\ \midrule
Europarl \textsc{En-Fr} & $1.7$  & $42.5$ \\ %
Europarl \textsc{En-De} & $1.7$  & $43.5$ \\ %
Giga \textsc{En-Fr}     & $18.3$ & $419.6$ \\ \bottomrule %
\end{tabular}
\caption{\label{tab:corpora}Training data size (in millions).}
\end{table}

\paragraph{Optimiser} For all architectures, we use the Adam optimiser \citep{Kingma+14:adam} with a learning rate of $10^{-3}$. 
Except where explicitly indicated, we
\begin{itemize}\setlength\itemsep{0em}
	\item train our models for $30$ epochs using mini batches of $100$ sentence pairs;
	\item use validation alignment error rate for model selection;
	\item train every model $10$ times with random \emph{Glorot} initialisation \citep{glorot2010initalization} and report mean and standard deviation;
	\item anneal the $\KL$ terms using the following schedule: we use a scalar $\alpha$ from $0$ to $1$ with additive steps of size $10^{-3}$ every $500$ updates. This means that at the beginning of the training, we allow the model to overfit to the likelihood terms, but towards the end we are optimising the true ELBO \citep{Bowman+16:conll}.
\end{itemize}
It is also important to highlight that we do not employ regularisation techniques (such as batch normalisation, dropout, or $L_2$ penalty) for they did not seem to yield consistent results. %

\subsection{\label{sec:wa}Word alignment}

Since our model leverages learning signal from parallel data by marginalising latent lexical alignments, we use alignment error rate to double check whether the model learns sensible word correspondences. 
Intrinsic assessment of word alignment quality requires manual annotation.
For English-French, we use the NAACL English-French hand-aligned data ($37$ sentence pairs for validation and $447$ for test) \citep{mihalcea2003aer}. 
For English-German, we use the data by \citet{padolapata2006endeAER} ($98$ sentence pairs for validation and $987$ for test).
Alignment quality is then measured in terms of alignment error rate (AER) \citep{Och+00:aer}---an F-measure over predicted alignment links. 
For prediction we condition on the posterior means $\mathbb E[Z_1^m]$ which is just the predicted variational means $\mathbf u_1^m$ and select the \Lx position for which $P(y_j, a_j|\mathbf u_1^m)$ is maximum (a form of approximate Viterbi alignment).

\begin{table}[h]\centering
\scalebox{0.83}{
\pgfplotstableread{results/europarl.en-fr.training.dev.txt}\loadedtable
\pgfplotstablecreatecol[
    create col/assign/.code={%
        \getthisrow{aer-mean}\mean
        \getthisrow{aer-std}\std
        \edef\entry{$\mean \pm \std$}%
        \pgfkeyslet{/pgfplots/table/create col/next content}\entry
    }	
]{AER}\loadedtable
\pgfplotstablecreatecol[
	create col/assign/.code={%
        \getthisrow{perp-x-mean}\mean
        \getthisrow{perp-x-std}\std
        \edef\entry{$\mean \pm \std$}%
        \pgfkeyslet{/pgfplots/table/create col/next content}\entry
    }
]{px}\loadedtable
\pgfplotstablecreatecol[
	create col/assign/.code={%
        \getthisrow{perp-y-mean}\mean
        \getthisrow{perp-y-std}\std
        \edef\entry{$\mean \pm \std$}%
        \pgfkeyslet{/pgfplots/table/create col/next content}\entry
    }
]{py}\loadedtable
\pgfplotstablecreatecol[
	create col/assign/.code={%
        \getthisrow{x-mean}\mean
        \getthisrow{x-std}\std
        \edef\entry{$\mean \pm \std$}%
        \pgfkeyslet{/pgfplots/table/create col/next content}\entry
    }
]{accx}\loadedtable
\pgfplotstablecreatecol[
	create col/assign/.code={%
        \getthisrow{y-mean}\mean
        \getthisrow{y-std}\std
        \edef\entry{$\mean \pm \std$}%
        \pgfkeyslet{/pgfplots/table/create col/next content}\entry
    }
]{accy}\loadedtable
\pgfplotstabletypeset[
	columns={config,accx,accy,AER},
	columns/config/.style={
		column name=Model,
		column type=l,
		string type
	},
	columns/accx/.style={
		column name=\Lx accuracy,
		string type
	},
	columns/accy/.style={
		column name=\Ly accuracy,
		string type
	},
	columns/AER/.style={
		column name=$\downarrow$AER,
		string type
	},
	every head row/.style={    
		before row=\toprule,
        after row=\midrule,
    },
	every last row/.style={
        after row=\bottomrule
	}
]\loadedtable
}
\caption{\label{tab:en-fr-dev}English-French validation $\uparrow$accuracy  and $\downarrow$ AER results.}
\end{table}

\begin{table}[h]
\centering
\scalebox{0.83}{
\pgfplotstableread{results/europarl.en-de.training.dev.txt}\loadedtable
\pgfplotstablecreatecol[
    create col/assign/.code={%
        \getthisrow{aer-mean}\mean
        \getthisrow{aer-std}\std
        \edef\entry{$\mean \pm \std$}%
        \pgfkeyslet{/pgfplots/table/create col/next content}\entry
    }	
]{AER}\loadedtable
\pgfplotstablecreatecol[
	create col/assign/.code={%
        \getthisrow{perp-x-mean}\mean
        \getthisrow{perp-x-std}\std
        \edef\entry{$\mean \pm \std$}%
        \pgfkeyslet{/pgfplots/table/create col/next content}\entry
    }
]{px}\loadedtable
\pgfplotstablecreatecol[
	create col/assign/.code={%
        \getthisrow{perp-y-mean}\mean
        \getthisrow{perp-y-std}\std
        \edef\entry{$\mean \pm \std$}%
        \pgfkeyslet{/pgfplots/table/create col/next content}\entry
    }
]{py}\loadedtable
\pgfplotstablecreatecol[
	create col/assign/.code={%
        \getthisrow{x-mean}\mean
        \getthisrow{x-std}\std
        \edef\entry{$\mean \pm \std$}%
        \pgfkeyslet{/pgfplots/table/create col/next content}\entry
    }
]{accx}\loadedtable
\pgfplotstablecreatecol[
	create col/assign/.code={%
        \getthisrow{y-mean}\mean
        \getthisrow{y-std}\std
        \edef\entry{$\mean \pm \std$}%
        \pgfkeyslet{/pgfplots/table/create col/next content}\entry
    }
]{accy}\loadedtable
\pgfplotstabletypeset[
	columns={config,accx,accy,AER},
	columns/config/.style={
		column name=Model,
		column type=l,
		string type
	},
	columns/accx/.style={
		column name=\Lx accuracy,
		string type
	},
	columns/accy/.style={
		column name=\Ly accuracy,
		string type
	},
	columns/AER/.style={
		column name=$\downarrow$AER,
		string type
	},
	every head row/.style={    
		before row=\toprule,
        after row=\midrule,
    },
	every last row/.style={
        after row=\bottomrule
	}
]\loadedtable
}
\caption{\label{tab:en-de-dev}English-German validation $\uparrow$accuracy and $\downarrow$AER results.}
\end{table}

We start by analysing validation results and selecting amongst a few variants of \textsc{EmbedAlign}. 
We investigate the use of annealing and the use of a bidirectional encoder in the variational approximation.   %
Table \ref{tab:en-fr-dev} (\ref{tab:en-de-dev}) lists $\downarrow$AER for \textsc{En-Fr} (\textsc{En-De}) as well as accuracy of word prediction.
It is clear that both annealing (systems decorated with subscript $\alpha$) and bidirectional representations improve the results across the board. 
In the rest of the paper we still investigate whether or not recurrent encoders help, but we always report results based on annealing.

In order to establish baselines for our models we report IBM models 1 and 2 \citep{Brown+1993:smt}. 
In a nutshell, IBM models 1 and 2 both estimate the conditional $P(y_j|x_1^m) = \sum_{a_j=1}^m P(a_j|m)P(y_j|x_{a_j})$ by marginalisation of latent lexical alignments. The only difference between the two models is the prior over alignments, which is uniform for IBM1 and categorical for IBM2. 
An important difference between IBM models and \textsc{EmbedAlign}  concerns the lexical distribution. 
IBM models are parameterised with independent categorical parameters, while our model instead is parameterised by a neural network. 
IBM models condition on a single categorical event $x_{a_j}$, namely, the word aligned to. Our model instead conditions on the latent embedding $z_{a_j}$ that stands for the word aligned to.

In order to establish even stronger conditional alignment models, we embed the conditioning words and replace IBM1's independent parameters by a neural network (single hidden layer MLP). We call this model a \emph{neural IBM1} (or \textsc{NIBM} for short).
Note that in an IBM model, the sequence $x_1^m$ is never modelled, therefore we can condition on it without restrictions. For that reason, we also experiment with a bidirectional LSTM encoder and condition lexical distributions on its hidden states.

\begin{table}[h]
\centering
\scalebox{0.94}{
\begin{tabular}{l c c}
\toprule
Model       & En-Fr & En-De  \\ \midrule
\textsc{IBM1}  & 32.45 & 46.71 \\  
\textsc{IBM2}  & 22.61 & 40.11 \\  
\textsc{NIBM}$_\text{BoW}$ & $27.35\pm0.19$   & $46.22\pm0.07$ \\
\textsc{NIBM}$_\text{BiRNN}$ & $25.57\pm0.40$   & $43.37\pm0.11$ \\
\textsc{EmbAlign}$_\text{BoW}$ & $30.97\pm2.53$ & $49.46\pm1.72$ \\
\textsc{EmbAlign}$_\text{BiRNN}$ & $29.43\pm1.84$ & $48.09\pm2.12$ \\ \bottomrule
\end{tabular}
}
\caption{\label{tab:test-aer}Test $\downarrow$AER.}
\label{my-label}
\end{table}

Table \ref{tab:test-aer} shows AER for test predictions. 
First observe that neural models outperform classic IBM1 by far, some of them even approach IBM2's performance. %
Next, observe that bidirectional encodings make \textsc{NIBM} much stronger at inducing good word-to-word correspondences.
\textsc{EmbedAlign} cannot catch up with \textsc{NIBM}, but that is not necessarily surprising.
Note that \textsc{NIBM} is a conditional model, thus it can use all of its capacity to better explain \Ly data. \textsc{EmbedAlign}, on the other hand, has to find a compromise between generating both streams of the data.
To make that point a bit more obvious, Table \ref{tab:acc-fr} (\ref{tab:acc-de}) lists accuracy of word prediction for \textsc{En-Fr} (\textsc{En-De}). 
Note that, without sacrificing \Ly accuracy, and sometimes even improving it,  \textsc{EmbedAlign} achieves very high \Lx accuracy. 
This still does not imply that induced representations have captured aspects of lexical semantics such as word senses.
All this means is that we have induced features that are jointly good at reconstructing both streams of the data one word at time.
Of course it is tempting to conclude that our models must be capturing some useful generalisations. 
For that, the next sections will investigate a range of semantic NLP tasks.

\begin{table}[h]
\centering 
\scalebox{0.95}{
\begin{tabular}{l c c}
\toprule
Model       &  \Lx accuracy & \Ly accuracy   \\ \midrule
\textsc{NIBM}$_\text{BoW}$ & - & $7.21\pm0.16$    \\
\textsc{NIBM}$_\text{BiRNN}$ & - & $6.47\pm0.45$    \\
\textsc{EmbAlign}$_\text{BoW}$ & $98.90\pm0.41$ & $7.08\pm0.34$  \\
\textsc{EmbAlign}$_\text{BiRNN}$ & $99.21\pm0.18$ & $7.44\pm0.61$ \\ \bottomrule
\end{tabular}
}
\caption{\label{tab:acc-fr}English-French $\uparrow$accuracy in test set}
\end{table}

\begin{table}[h]
\centering 
\scalebox{0.95}{
\begin{tabular}{l c c}
\toprule
Model       &  \Lx accuracy & \Ly accuracy   \\ \midrule
\textsc{NIBM}$_\text{BoW}$ & - & $7.94\pm0.03$    \\
\textsc{NIBM}$_\text{BiRNN}$ & - & $8.38\pm0.10$    \\
\textsc{EmbAlign}$_\text{BoW}$ & $96.86\pm2.89$ & $8.72\pm0.39$  \\
\textsc{EmbAlign}$_\text{BiRNN}$ & $99.32\pm0.34$ & $8.00\pm0.12$ \\ \bottomrule
\end{tabular}
}
\caption{\label{tab:acc-de}English-German $\uparrow$accuracy in test set}
\end{table}

\begin{table}
\centering
\begin{tabular}{l c c}
\toprule
Model                                                           & $\cos$ & $\KL$ \\ \midrule
\textsc{Random}   & $30.0$ & - \\
\textsc{SkipGram} & $44.9$ & - \\ 
\textsc{BSG}      & -      & $46.1$ \\ \hline
\textsc{En}$_\text{BoW}$ & $29.75\pm0.55$ & $27.93\pm0.25$\\
\textsc{En}$_\text{BiRNN}$ & $21.31\pm1.05$ & $27.64\pm0.40$ \\ \hline
\textsc{En-Fr}$_\text{BoW}$ & $42.72\pm0.36$ & $41.90\pm0.35$ \\
\textsc{En-Fr}$_\text{BiRNN}$ & $42.19\pm0.57$ & $41.61\pm0.55$ \\
\textsc{En-De}$_\text{BoW}$ & $41.90\pm0.58$ & $40.63\pm0.55$ \\
\textsc{En-De}$_\text{BiRNN}$ & $42.07\pm0.47$ & $40.93\pm0.59$ \\ \bottomrule
\end{tabular}
\caption{English $\uparrow$GAP on LST test data.}
\label{tab:gap}
\end{table}

\subsection{Lexical substitution task}

The English lexical substitution task (LST) consists in selecting a substitute word for a target word \emph{in context} \citep{McCarthy+2009:LST}. 
In the most traditional variant of the task, systems are presented with a list of potential candidates and this list must be sorted by relatedness.

\paragraph{Dataset} The LST dataset includes $201$ target words present in $10$ sentences/contexts each, along with a manually annotated list of potential replacements. The data are split in $300$ instances for validation and $1,710$ for test. Systems are evaluated by comparing the predicted ranking to the manual one in terms of generalised average precision (GAP) \citep{Melamud+15:naacl}. 

\paragraph{Prediction} We use \textsc{EmbedAlign} to encode each candidate (in context) as a posterior Gaussian density. Note that this task dispenses with inferences about \Ly. 
Each candidate is compared to the target word in context through a measure of overlap between their inferred densities---we take $\KL$ divergence.
We then rank candidates using this measure.

\begin{table*}[t]
\centering
\scalebox{0.95}{
\begin{tabular}{lcccccccccc}
\toprule
Model    & MR    & CR    & SUBJ  & MPQA  & SST  & TREC & MRPC        & SICK-R & SICK-E & SST14     \\ \midrule
\textsc{w2vec} & 77.7  & 79.8  & 90.9  & 88.3  & 79.7 & 83.6 & 72.5/81.4   & 0.80  & 78.7   & 0.65/0.64 \\
\textsc{NMT} & 64.7  & 70.1  & 84.9  & 81.5  & - & 82.8 & -/-   & -  & -   & 0.43/0.42 \\ \hline
\textsc{En}   & 57.6  & 66.2  &  70.9   & 71.8  & 58.0   & 62.9   & 70.3/80.1   & 0.62  & 73.7   & 0.54/0.55 \\
\textsc{En-Fr}    & 63.5  & 71.5  &  78.9   &  82.3 & 65.1   & 62.1   & 71.4/80.5   & 0.69  &  75.9  & 0.69/0.59  \\
\textsc{En-De}    & 64.0  & 68.9  &  77.9   & 81.8  & 65.1   & 59.5   & 71.2/80.5   &  0.69 &  74.8  & 0.62/0.61  \\
\textsc{Combo} & 66.7  &  73.1 & 82.4    & 84.8  & 69.2   & 67.7   & 71.8/80.7   & 0.73  & 77.4   & 0.62/0.61  \\ \bottomrule
\end{tabular}
}
\caption{English sentence evaluation results: the last four rows correspond to the mean of 10 runs with \textsc{EmbedAlign} models. All models, but \textsc{w2vec}, employ bidirectional encoders.}
\label{tab:senteval}
\end{table*}

Table \ref{tab:gap} lists GAP scores for variants of \textsc{EmbedAlin} (bottom section) as well as some baselines and other established methods (top section).
For comparison, we also compute GAP by sorting candidates in terms of cosine similarity, in which case we take the Gaussian mean as a summary of the density.
The top section of the table contains systems reported by \citet{Melamud+15:naacl} (\textsc{Random} and \textsc{SkipGram}) and by \citet{Brazinskas2017} (\textsc{BSG}).
Note that both \textsc{SkipGram} \citep{Mikolov+2013:DR} and \textsc{BSG} were trained on the very large ukWaC English corpus \citep{Ferraresi08introducingand}. 
\textsc{SkipGram} is known to perform remarkably well regardless of its apparent insensitivity to context (in terms of design). 
\textsc{BSG} is a close relative of our model which gives \textsc{SkipGram} a Bayesian treatment (also by means of amortised variational inference) and is by design sensitive to context in a manner similar to \textsc{EmbedAlign}, that is, through its inferred posteriors. 

Our first observation is that cosine seems to outperform $\KL$ slightly.
Others have shown that $\KL$ can be used to predict directional entailment \citep{vilnis2014word,Brazinskas2017}, since LST is closer to paraphrasing than to entailment directionality may be a distractor, but we leave it as a rather speculative point.
One additional point worth highlighting: the middle section of Table \ref{tab:gap}.
\textsc{En}$_\text{BoW}$ and \textsc{En}$_\text{BiRNN}$ show what happens when we do not give \textsc{EmbedAlign} \Ly supervision at training. 
That is, imagine the model of Figure \ref{fig:model} without the bottom plate.
In that case, the model representations overfit for \Lx word-by-word prediction. Without the need to predict any notion of context (monolingual or otherwise), the representations drift away from semantic-driven generalisations and fail at lexical substitution. %

\subsection{Sentence Evaluation}

\citet{conneau2017supervised} developed a framework to evaluate unsupervised sentence level representations trained on large amounts of data on a range of supervised NLP tasks. 
We assess our induced representations using their framework on the following benchmarks evaluated on classification $\uparrow$accuracy (MRPC is further evaluated on $\uparrow$F1)
\begin{description}\setlength\itemsep{0em}
\item[MR] classification of positive or negative movie reviews;
\item[SST] fined-grained labelling of movie reviews from the Stanford sentiment treebank (SST);
\item[TREC] classification of questions into $k$-classes;
\item[CR] classification of positive or negative product reviews;
\item[SUBJ] classification of a sentence into subjective or objective;
\item[MPQA] classification of opinion polarity;
\item[SICK-E] textual entailment classification;
\item[MRPC] paraphrase identification in the Microsoft paraphrase corpus;
\end{description}
as well as the following benchmarks evaluated on the indicated correlation metric(s)
\begin{description}\setlength\itemsep{0em}
\item[SICK-R] semantic relatedness between two sentences ($\uparrow$Pearson);
\item[SST-14] semantic textual similarity ($\uparrow$Pearson/Spearman).
\end{description}

\paragraph{Prediction} We use \textsc{EmbedAlign} to annotate every word in the training set of the benchmarks above with the posterior mean embedding in context. 
We then average embeddings in a sentence and give that as features to a logistic regression classifier trained with $5$-fold cross validation.\footnote{\url{http://scikit-learn.org/stable/}} 

For comparison, we report a \textsc{SkipGram} model (here indicated as \textsc{w2vec}) as well as a model that uses the encoder of a neural machine translation system (NMT) trained on English-French Europarl data. In both cases, we report results by \citet{conneau2017supervised}. 
Table \ref{tab:senteval} shows the results for all benchmarks.\footnote{In Appendix \ref{app:runs} we provide bar plots marked with error bars ($2$ standard deviations).} 
We report \textsc{EmbedAlign} trained on either \textsc{En-Fr} or \textsc{En-De}.
The last line (\textsc{Combo}) shows what happens if we train logistic regression on the concatenation of embeddings inferred by both \textsc{EmbedAlign} models, that is, \textsc{En-Fr} and \textsc{En-De}.
Note that these two systems perform sometimes better sometimes worse depending on the benchmark.
There is no clear pattern, but differences may well come from some qualitative difference in the induced latent space.
It is a known fact that different languages realise lexical ambiguities differently, thus representations induced towards different languages are likely to capture different generalisations.\footnote{We also acknowledge that our treatment of German is likely suboptimal due to the lack of subword features, as it can also be seen in AER results.}
As \textsc{Combo} results show, the representations induced from different corpora are somewhat complementary.
That same observation has guided paraphrasing models based on pivoting \citep{BannardEtAl2005P}.
Once more we report a monolingual variant of \textsc{EmbedAlign} (indicated by \textsc{En}) in an attempt to illustrate how crucial the translation signal is.

\subsection{\label{sec:wordsim}Word similarity}

Word similarity benchmarks are composed of word pairs which are manually ranked out of context.
For completeness, we also tried evaluating our embeddings in such benchmarks despite our work being focussed on applications where context matters.

\paragraph{Prediction} To assign an embedding for a word type, we infer Gaussian posteriors for all training instances of that type in context and aggregate the posterior means through an average (effectively collapsing all instances). 

To cover the vocabulary of the typical benchmark, we have to use a much larger bilingual collection than Europarl. Based on the results of \S\ref{sec:wa}, we decided to proceed with English-French only---recall that models based on that pair performed better in terms of AER. 
Results in this section are based on \textsc{EmbedAlign} (with bidirectional variational encoder) trained on the Giga web corpus (see Table \ref{tab:corpora} for statistics). Due to the scale of the experiment, we report on a single run.

We trained on Giga with the same hyperparameters that we trained on Europarl, however, for $3$ epochs instead of $30$ (with this dataset an epoch amounts to $183,000$ updates). 
Again, we performed model selection on AER. %
Table \ref{tab:wordsim} shows the results for several datasets using the framework of \citet{faruquiEtal2014WSim}. 
Note that \textsc{EmbedAlign} was designed to make use of context information, thus this evaluation setup is a bit unnatural for our model. Still, it outperforms \textsc{SkipGram} on 5 out of 13 benchmarks, in particular, on SIMLEX-999, whose relevance has been argued by  \citet{upadhyay-EtAl:2016:P16-1}. 
We also remark that this model achieves $0.25$ test AER and $45.16$ test GAP on lexical substitution---a considerable improvement compared to models trained on Europarl and reported in Tables \ref{tab:test-aer} (AER) and \ref{tab:gap} (GAP). 

\begin{table}\centering
\begin{tabular}{l r r }
\toprule
Dataset &  {\small \textsc{SkipGram}}  &  {\small \textsc{EmbedAlign}} \\ \midrule
    MTurk-771   &  0.5679  & 0.5229  \\
    SIMLEX-999  &  0.3131  & 0.3887  \\
    WS-353-ALL  &  0.6392  & 0.3968  \\
        YP-130  &  0.3992  & 0.4784  \\
      VERB-143  &  0.2728  & 0.4593  \\
     MEN-TR-3k  &  0.6462  & 0.4191  \\
  SimVerb-3500  &  0.2172  & 0.3539  \\
         RG-65  &  0.5384  & 0.6389  \\
    WS-353-SIM  &  0.6962  & 0.4509  \\
   RW-STANFORD  &   0.3878 & 0.3278  \\
    WS-353-REL  &  0.6094  & 0.3494  \\
         MC-30  &  0.6258  & 0.5559  \\
     MTurk-287  &  0.6698  & 0.3965  \\ \bottomrule
\end{tabular}
\caption{\label{tab:wordsim}Evaluation of English word embeddings out of context in terms of Spearman's rank correlation coefficient ($\uparrow$). The first column is from \citep{faruquiEtal2014WSim}.}

\end{table}

\section{\label{sec:related}Related work} 

Our model is inspired by lexical alignment models such as IBM1 \citep{Brown+1993:smt}, however, we generate words $y_1^n$ from a latent vector representation $z_1^m$ of $x_1^m$, rather than directly from the observation $x_1^m$.
IBM1 takes \Lx sequences as conditioning context and does not model their distribution. %
Instead, we propose a joint model, where \Lx sentences are generated from latent embeddings.

There is a vast literature on exploiting multilingual context to strengthen the notion of synonymy captured by monolingual models.
Roughly, the literature splits into two groups, namely, approaches that derive additional features and/or training objectives based on pre-trained alignments  \citep{klementiev-titov-bhattarai:2012:PAPERS,faruqui-dyer:2014:EACL,luong2015bilingual,VsusterEtAl2016MS}, and approaches that promote a joint embedding space by working with sentence level representations that dispense with explicit alignments \citep{hermann-blunsom:2014:P14-1,ap2014autoencoder,BilBOWA,hill2014embedding}.

The work of \citet{Kocisky+14:vecibm2} is closer to ours in that they also learn embeddings by marginalising alignments, however, their model is conditional---much like IBM models---and their embeddings are not part of the probabilistic model, but rather part of the architecture design. %
The joint formulation allows our latent embeddings to harvest learning signal from \Ly while still being driven by the learning signal from \Lx---in a conditional model the representations can become specific to alignment deviating from the purpose of well representing the original language. In \S\ref{sec:exp} we show substantial evidence that our model performs better when using both learning signals.

\citet{vilnis2014word} first propose to map words into Gaussian densities instead of point estimates for better word representation. %
For example, a distribution can capture asymmetric relations that a point estimate cannot. 
\citet{Brazinskas2017} recast the skip-gram model as a conditional variational auto-encoder.  They induce a Gaussian density for each occurrence of a word in context, and for that their model is the closest to ours, but training is based on prediction of neighbouring words.
Unlike our model, the Bayesian skip-gram still requires generation of negative samples for discriminative training. 
It is perhaps worth highlighting that, in principle, both strategies can be combined.

\section{Discussion}

We have presented a generative model of word representation that learns from positive correlations implicitly expressed in translation data.
In order to make these correlations surface, we induce and marginalise latent lexical alignments.

Embedding models such as CBOW and skip-gram \citep{Mikolov+2013:DR} are essentially speaking supervised classifiers.
This means they depend on somewhat artificial strategies to derive labelled data from monolingual corpora---words far from the central word still have co-occurred with it even though they are taken as negative evidence.
Training our proposed model does not require a heuristic notion of negative training data. However, the model is also based on a somewhat artificial assumption: \Lx words do not necessarily need to have an \Ly equivalent and, even when they do, this equivalent need not be realised as a single word. 

We have shown with extensive experiments that our model can induce representations useful to several tasks including but not limited to alignment (the task it most obviously relates to). 
We observed interesting results on semantic natural language processing benchmarks such as natural language inference, lexical substitution, paraphrasing, and sentiment classification.

We are currently expanding the notion of distributional context to multiple auxiliary foreign languages at once. 
This seems to only require minor changes to the generative story and could increase the model's disambiguation power dramatically. 
Another direction worth exploring is to extend the model's hierarchy with respect to how parallel sentences are generated. 
For example, modelling sentence level latent variables may capture global constraints and expose additional correlations to the model.

\section*{Acknowledgments}

We thank Philip Schulz for comments on an earlier version of this paper as well as the anonymous NAACL reviewers. 
One of the Titan Xp cards used for this research was donated by the NVIDIA Corporation.
This work was supported by the Dutch Organization for Scientific Research (NWO) VICI Grant nr. 277-89-002.

\bibliography{BIB}
\bibliographystyle{acl_natbib}

\appendix
\clearpage
\section{\label{app:runs}Multiple runs sentence evaluation}
\begin{figure}[h]	
	\begin{center}
		\includegraphics[scale=0.56]{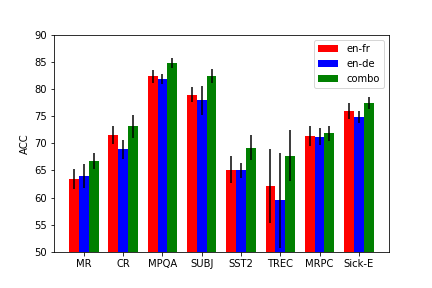}
		\includegraphics[scale=0.56]{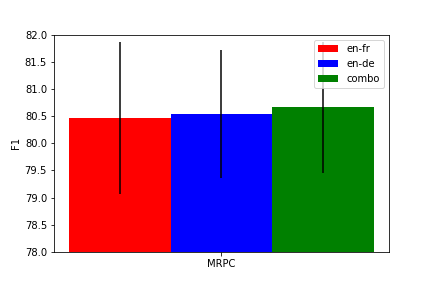}
		\includegraphics[scale=0.56]{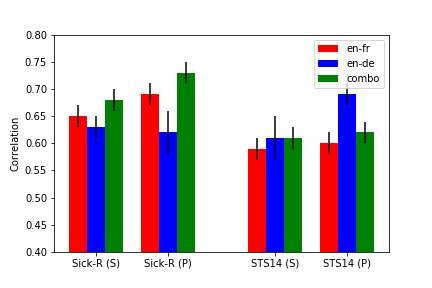}
	\end{center}
	\caption{\label{fig:mult} Mean and two standard deviations (error bars) for 10 runs of \textsc{EmbedAlign} on the sentence evaluation benchmarks.}
\end{figure}

Figure \ref{fig:mult} shows multiple runs of our proposed model on sentence evaluation. The first figure reports the mean and two standard deviations (error bars) for benchmarks based on accuracy (ACC), the second figure reports benchmarks based on F1, and finally the third figure reports benchmarks based on correlation metrics Spearman (S) and Pearson (P).

\section{\label{app:architecture}Architecture}

\begin{figure}[h]	
	\begin{center}
		\includegraphics[scale=0.7]{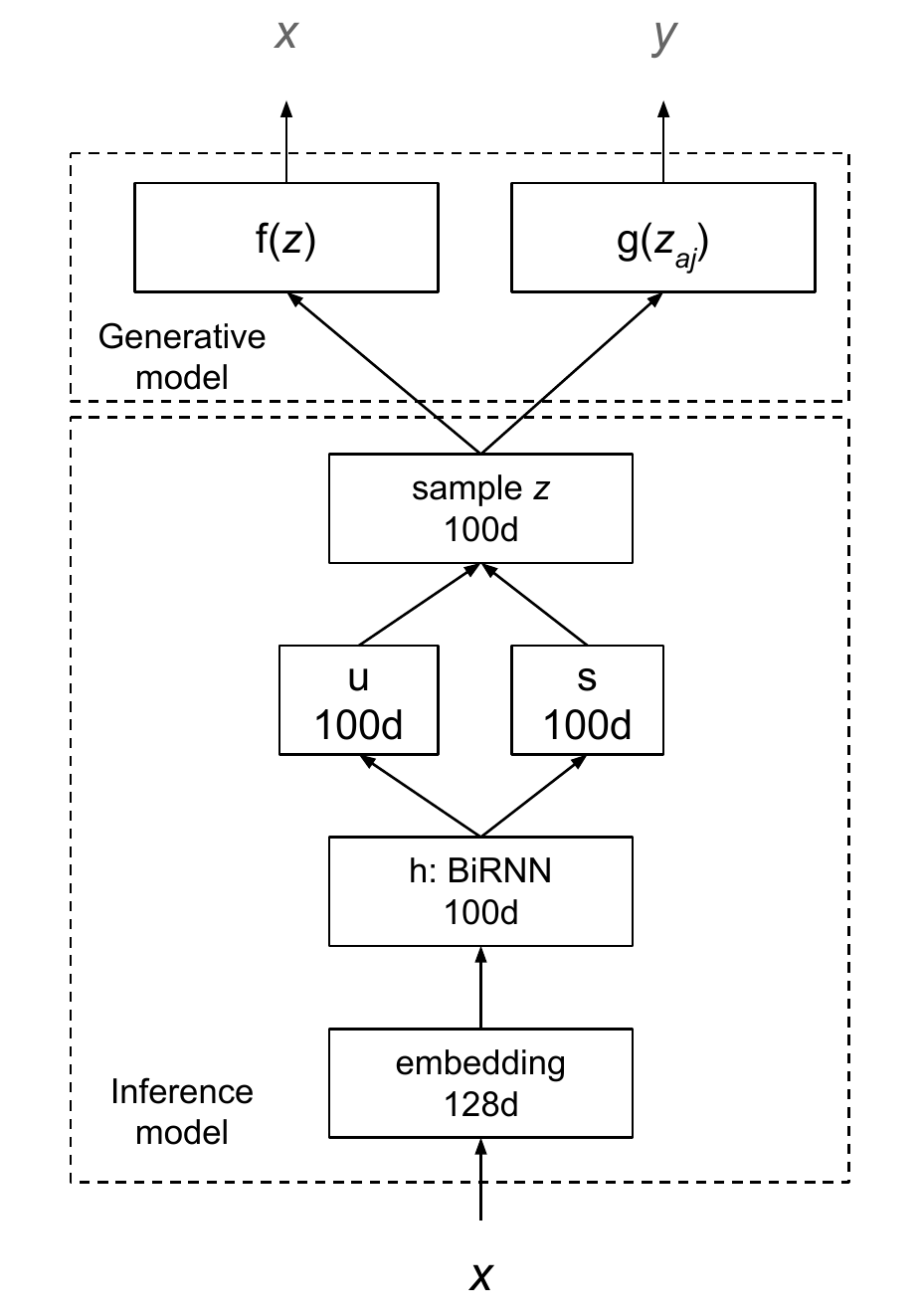}
	\end{center}
	\caption{\label{fig:arch}  Architecture for EmbedAlign.}
\end{figure}

Figure \ref{fig:arch} shows the architecture for the inference and generative models in \textsc{EmbedAlign}, with BiRNN encoder ($h$).

\end{document}